\documentclass{article}

\usepackage{PRIMEarxiv}
\usepackage[utf8]{inputenc} 
\usepackage[T1]{fontenc}    
\usepackage{hyperref}       
\usepackage{url}            
\usepackage{booktabs}       
\usepackage{amsfonts}       
\usepackage{nicefrac}       
\usepackage{microtype}      
\usepackage{lipsum}
\usepackage{fancyhdr}       
\usepackage{graphicx}       
\graphicspath{{media/}}     
\usepackage{amsmath}
\usepackage{todonotes}

\usepackage{listings}
\title{Crowd Score: A Method for the Evaluation of Jokes using Large Language Model AI Voters as Judges}

\author{
  Fabrício Góes, Zisen Zhou \\
  Computing and Mathematical Sciences Department \\
  University of Leicester \\
  Leicester, UK\\
  \texttt{\{fabricio.goes,zz254\}@leicester.ac.uk},
  \And
  Piotr Sawicki, Marek Grzes \\
  School of Computing \\
  University of Kent \\
  Canterbury, UK\\
  \texttt{\{p.sawicki,m.grzes\}@kent.ac.uk} \\
  \And
  Daniel G. Brown \\
  David R. Cheriton School of Computer Science
  University of Waterloo \\
  Waterloo, Canada\\
  \texttt{dan.brown@uwaterloo.ca}
}

\begin{document}
\maketitle

\begin{abstract}
This paper presents the Crowd Score, a novel method to assess the funniness of jokes using large language models (LLMs) as AI judges. Our method relies on inducing different personalities into the LLM and aggregating the votes of the AI judges into a single score to rate jokes. We validate the votes using an auditing technique that checks if the explanation for a particular vote is reasonable using the LLM. We tested our methodology on 52 jokes in a crowd of four AI voters with different humour types: affiliative, self-enhancing, aggressive and self-defeating. Our results show that few-shot prompting leads to better results than zero-shot for the voting question. Personality induction showed that aggressive and self-defeating voters are significantly more inclined to find more jokes funny of a set of aggressive/self-defeating jokes than the affiliative and self-enhancing voters. The Crowd Score follows the same trend as human judges by assigning higher scores to jokes that are also considered funnier by human judges. We believe that our methodology could be applied to other creative domains such as story, poetry, slogans, etc. It could both help the adoption of a flexible and accurate standard approach to compare different work in the CC community under a common metric and by minimizing human participation in assessing creative artefacts, it could accelerate the prototyping of creative artefacts and reduce the cost of hiring human participants to rate creative artefacts. \footnote{Abstract fully written by GPT-3 using the introduction and related work sections as input.}
\end{abstract}

\keywords{Large Language Models \and Jokes Evaluation \and Computational Creativity \and Crowd Score \and Prompt Engineering \and AI judges \and Personality Induction}

\section{Introduction}
In Computational Creativity (CC), there are two main strategies to assess the creativity of artefacts: evaluation metrics and human judges \cite{jordanous2012,franca2016}. Evaluation metrics, proposed by humans, are usually used by generative systems to evaluate novelty and value of potential creative artefact candidates. The best ones are ultimately evaluated by humans, since they are still the ultimate judges on creativity. Despite evidence that non-expert judges cannot appropriately evaluate the creativity of a human or machine \cite{lamb2015}, studies have relied on hiring non-expert volunteers on crowd sourcing platforms, such as Amazon Mechanical Turk, to evaluate/rate artifacts in the creative domain \cite{toplyn2022}. These studies usually do not ask volunteers to explain the reasoning behind their scores, but accept their judgement as valid.
In this paper, we assume that machines, much like non-expert humans, can be used to judge the creativity of artefacts. This assumption is backed on recent advances of large language models (LLMs) such as GPT-3 \cite{brown2020}, that enable emergent behaviour through few-shot and augmenting prompting \cite{wei2022,hessel2022}, that is, abilities that were not present in smaller models. In particular, jokes and humour are challenging for machines because they involve complex concepts such as irony, sarcasm, and puns \cite{veale2022}. Jokes often rely on cultural context and knowledge, which can be difficult for a machine to access. However, recent work \cite{wang2022,hessel2022,shatnawi2022,tian2022,mitall2022,liu2021,jiang2022} shows that prompting or fine-tuning LLMs for humour detection is a viable approach. On top of it, some recent publications show that LLMs can be configured/prompted to assume different personalities with zero and few-shot prompting \cite{kojima2022}. This favours the creation of a crowd of AI voters, where each one's vote is aggregated into a rating/score that can accurately measure the level of funniness of jokes, the CC-Crowd Score. In order to validate those votes, we apply an auditing technique that checks if the explanation for a particular vote is reasonable using the LLM.
We believe that this method could be applied to other creative domains such as story, poetry, slogans, etc. It could both help the adoption of a flexible and accurate standard approach to compare different work in the CC community under a common metric and by minimizing human participation in assessing creative artefacts, we can accelerate the prototyping of creative artefacts and reduce the cost of hiring human participants to rate creative artefacts. In this paper, we focus on evaluating funniness of jokes as a case study.

We tested our methodology to assess the funniness of jokes from \cite{toplyn2022} in a crowd of four voters with different humour types \cite{martin2003}. Our results show that: i) few-shot prompting leads to better results than zero-shot for the voting question, where picking the least appropriate opposite word can reduce balanced accuracy by 26\% and 25\% for zero-shot and few-shot respectively; ii) personality induction showed that aggressive and self-defeating voters are significantly more inclined to find more jokes funny of a set of aggressive/self-defeating jokes than the affiliative and self-enhancing voters; and iii) the Crowd Score follows the same trend as human judges by assigning higher scores to jokes that are also considered funnier by human judges.

Our main contributions are: 
\begin{itemize}
    \item The Crowd Score, a novel method to assess jokes with LLMs using their intrinsic evaluation metrics. It relies on AI voters as judges for creativity instead of human judges, in which a crowd of AI voters are induced in a LLM and their votes are aggregated into a single score to rate jokes.
    \item An auditing technique to validate the votes of the AI judges using LLMs.
    \item A case study with 52 jokes and 4 induced personalities to assess the funniness of jokes.
    \item A set of prompt templates that could be customized to assess other creative artefacts.
\end{itemize}

The rest of this paper is organized as follows. In Section \ref{related_work}, we present the background and recent related work to support this research. In Section \ref{crowd_score}, we describe our Crowd Score method in details. Section \ref{results} presents and analyzes the experimental results. This paper is concluded in Section \ref{conclusion}.

\section{Related Work}\label{related_work}

A major issue on humour research evaluation has been the lack of comprehensive datasets of jokes with ratings of funniness. \cite{hossain2020} points out that most humour datasets are usually annotated in a binary fashion (funny or not funny), which does not capture the level of funniness. A fine grained evaluation of jokes reduces the chance of misjudging jokes that are considered borderline on average, but funny for particular groups of people. However, some recent publications make available databases with annotations on the rating of jokes \cite{toplyn2022,sun2022}. \cite{hossain2020} also tackles that issue using crowd-sourcing to ask human volunteers to rate the funniness of modified headlines. 

The use of large language models (LLMs), such as GPT-3, is becoming more prevalent for generating humorous texts. For instance, \cite{wang2022} rely on human participants to evaluate Chinese crosstalks (comic dialogues) generated by a fine-tuned GPT-3 in regard to general quality, humour, coherence and ethical risky content. The results show that the best generation achieved 65\% of general quality, and that the humour criterion is not satisfied. It is important to remark that they used the standard BLEU metric to evaluate the performance of GPT-3. \cite{mitall2022} also generates puns using GPT-3. Based on context words with ambiguous meanings, they provide the target pun word and its two meanings to GPT-3 and prompt it to generate puns. They also ask humans to evaluate if the generated puns are indeed puns and rate how funny they are. In \cite{tian2022}, the authors propose a pun generator which, given a pun word pair, retrieves a context word and phrase, and they use GPT-2 to produce a pun. Human evaluators judged that their method achieved the highest funniness among other pun generators. In \cite{shatnawi2022}, the authors propose an approach to predict the funniness of news headlines using the BERT model. The approach was tested with the humour datasets from the SemEval-2020 workshop, achieving high performance.  We can observe that there are two main approaches to LLM's: fine-tuning \cite{wang2022,hessel2022,shatnawi2022,tian2022} and prompting \cite{mitall2022,liu2021,jiang2022}. However, most related work relies on human evaluators as the final judges of humour. We depart from this practice and claim that LLMs can be used as humour evaluators, even in their current stage.

An innovative approach is described in \cite{hessel2022}, in which the authors evaluate if large language models are capable of evaluating and explaining captions of the New Yorker Caption Contest. Those captions are humourous sentences describing a cartoon. Results show that a fine-tuned GPT-3 cannot recognize the captions' relevance, evaluate or explain them as effectively as humans. However, their partial capacity can be sufficient to work as creative collaborators. This work strongly corroborates our claim that despite being imperfect, current LLMs can be used as judges of humour. We move one step further and rely on concepts introduced in \cite{jiang2022} to induce LLMs with different personalities to provide different opinions/rating about the humour of a joke.

In \cite{jiang2022}, the authors propose a method to induce a certain personality on large language models. The personality is based on the Big Five personality factors. The paper compares two approaches. The naive one uses zero-shot learning to prompt a personality using ``You are a/an X person'', where X is one of the five factors, and the ``Word Level Auto-Prompt'' approach imposes the 3 most significant words to describe each factor. Both approaches are evaluated by prompting the model to answer the Big Five questionnaire. Results show that it is possible to simulate the desired personality using both methods. Another work \cite{aher2022} uses GPT-3 to simulate responses of humans by varying their names and other details, under some human experiments such as the Ultimatum Game. Results show that GPT-3 responses are consistent with prior human studies. Another study \cite{meyer2022} used GPT-3 to generate data to train conversational models and compared them with real data generated by humans. The authors assume that large language models can be used to replace humans when data is scarce. In our proposal, we combine the possibility of inducing a certain personality with the capability of LLMs to evaluate the humour level of jokes to create a method, as an alternative to traditional metrics, that could also be applied to evaluate jokes.

The use of LLMs as an alternative to traditional metrics is supported by recent research in \cite{goyal2022}, where the authors show that automatic reference-free and reference-based metrics, such as BLEU, BERTScore, BLANC and QuestEval, are ineffective to evaluate the quality of news summaries  generated by zero-shot GPT-3, when compared to the human evaluation. Instead of relying on a single metric, we use the idea of aggregation as in \cite{arora2022}, where aggregation is used to Q\&A tasks for large language models. By using prompt chains, an input claim is converted into a question, and multiple noisy answers are generated. Those answers, which are binary (yes/no), are aggregated using weak supervision into a final prediction. Our work proposes aggregation of binary votes generated by multiple voters with induced personalities in large language models, instead of imperfect/noisy answers as in \cite{arora2022}.

Another issue is the reliability of prompting GPT-3. In \cite{si2022}, prompts are created to induce reliable behaviors on GPT-3. The paper shows that retrieving evidence passages for Q\&A problems can improve the performance of GPT-3. It also shows that GPT-3 memorized answers can be updated by adding conflicting passages in the prompts, allowing it to output different answers in accordance with this new context. Differently, in our work, we employ a voting question prompt to ensure that we pick the best pair of words in the binary classification/voting. We also propose the concept of auditing, where the LLM is used to check if the explanations of the votes by each voter are reasonable/consistent before computing the respective votes as valid. 

Finally, in terms of creativity, on top of the other related work presented, \cite{stevenson2022} assessed GPT-3's creativity using the Guilford's Alternative Uses Test and compared with human responses on value, novelty and surprise. They have used a very detailed, handcrafted prompt to instruct GPT-3 to list the creative uses of objects. Their results show that humans rating outperform GPT-3 only by a small difference. This is yet another evidence that endorses the current capacity of GPT-3 to address problems in the creative domain. 

In this paper, we argue that by deploying a large language model equipped with voters with different personalities, we can accurately rate the funniness of jokes, without the need for human judges. 

\section{Crowd Score}\label{crowd_score}

The main goal of the Crowd Score method is to provide a method to assess the creativity of artifacts using a crowd of AI as judges. This method consists of the following steps: i) Voting Question, ii) Personality Induction, iii) Auditing and iv) Score Aggregation. First, a voting question that will be prompted to the AI crowd should be selected. Secondly, each AI voter should be configured to a different personality. Those personalities should reflect the audience with the appropriate traits that are relevant to the assessment of the creative artifact. For example, in the jokes domain, the humour type is the most important trait. Then, each personality should reply to the voting question with an answer (e.g. funny or not funny) and an explanation about the reasoning behind this vote. This is validated by an auditing prompt to ensure that votes are based on solid/reasonable argumentation, instead of randomness. The final step is to aggregate the individual votes to form the crowd score that indicates how much creative is an artifact. Those steps will be detailed in the following sections. 

These sections are also accompanied by a set of prompts to implement those steps. We use the following notation for prompting. Slots are equivalent to variables, where their content can be stored or updated. Slots are in the form: "Identifier: \$Description", where Identifier is the name of the slot and \$Description is the content of a slot. For example: "Joke: Why did the chicken run across the road? To get to the other side.". The notation [Identifier] is used to reference a particular slot, for example, the following prompt displays the content of a slot: "Show the content of [Joke].".

\subsection{Voting Question}

This first step consists of identifying the most appropriate prompt for the voting question. This prompt should be crafted in a way that leads to high accuracy in the LLM prediction for a subset of known positives and negatives, respectively funny and not funny jokes in this research. This is important, as we show in Section \ref{results}, because depending on the pair of words used, the LLM accuracy can vary significantly. For jokes, the prompt should make the LLM model decide if a joke is funny or not funny. Figures \ref{prompt_voting_zero} and \ref{prompt_voting_few} show zero-shot and few-shot prompts. In the few-shot prompt, examples of a funny joke and a not funny one are provided. In our experiments, we fixed the "funny" word and varied the \$Opposite (e.g. not funny, dumb, boring, etc.) until we found the opposite word that achieves the best accuracy in identifying positives and negatives. 

\begin{figure}
\begin{footnotesize}
\begin{lstlisting}
Classify the following [Joke] as Funny or $Opposite.

Joke: $JokeDescription
Classification:
\end{lstlisting}
\end{footnotesize}
\caption{Zero-shot voting question to classify the joke as funny or its opposite (not funny, dumb, boring etc.).}
\label{prompt_voting_zero} 
\end{figure}

\begin{figure}
\begin{footnotesize}
\begin{lstlisting}
Classify the following [Joke] as Funny or $Opposite.

Joke: $FunnyJokeDescription
Classification: Funny.

Joke: $NotFunnyJokeDescription
Classification: $Opposite.

Joke: $JokeDescription
Classification:
\end{lstlisting}
\end{footnotesize}
\caption{Few-shot voting question prompt to classify the joke as funny or its opposite (not funny, dumb, boring etc.).}
\label{prompt_voting_few} 
\end{figure}

\subsection{Personality Induction}

Once the voting question prompt is tuned to achieve high accuracy, our method requires the definition of the shortest and the most accurate set of traits that should be used to describe each personality \cite{jiang2022}. Zero-shot prompting can be used if the LLM model has knowledge about the personality traits. In particular, the classification of jokes there are four types of humour: affiliative, self-enhancing, aggressive and self-defeating \cite{martin2003}. Affiliative humor is a non-hostile, tolerant use of humor that is affirming of self and others. Self-enhancing humor is used to make people feel good about themselves. Aggressive humor is the use of sarcasm, teasing, ridicule, derision, and put-downs. Finally, self-defeating humor involves poking fun at oneself for the enjoyment of others. The LLM was prompted to define those types of humour and they were described correctly, so in this research we assume that zero-shot prompt is enough to induce personalities with those traits. Since the type of humour is the only important trait in our experiments, we create personalities induced by the following prompt: "Classify the following [Joke] as Funny or \$Opposite \textbf{as a person that enjoys \$TypeOfHumour humour.}". The full prompt can be seen in Figure \ref{prompt_personality}. 

The use of different personalities is crucial to the creation of a diverse crowd of AI voters. Depending on the creative domain, target public and LLM, the number of personalities and traits can vary significantly and zero-shot might not be possible. Instead, it might be necessary to describe the personalities from scratch and with enough detail.

\begin{figure}
\begin{footnotesize}
\begin{lstlisting}
Classify the following [Joke] as Funny or $Opposite as a person that enjoys $TypeOfHumour humour.

Joke: $JokeDescription
Classification:
\end{lstlisting}
\end{footnotesize}
\caption{Prompt for personality induction.}
\label{prompt_personality} 
\end{figure}

\subsection{Auditing}
Each vote of each personality needs to be validated before it is accepted and included as part of the crowd score. Reliability is an issue with current LLMs, and additional checks should be conducted to assess if the LLM outputted a vote based on a logical reasoning and not just as noise from the model. In order to achieve that goal, two prompts were designed. The first one prompts the LLM for a reasoning using the CoT (Chain-of-Thought) prompt ``Let's think step by step'' extensively used in the recent literature \cite{zhang2022,kojima2022,arora2022}. This powerful prompt forces the LLM to produce a step-by-step explanation of the reasoning behind the output answer, in our case, why the joke is funny or not. The full prompt is shown in Figure \ref{prompt_explanation}.

\begin{figure}
\begin{footnotesize}
\begin{lstlisting}
Classify the following [Joke] as Funny or $Opposite as a person that enjoys $TypeOfHumour humour.

Joke: $JokeDescription
Classification: $FunnyOrOpposite. Let's think step by step why this [Joke] is $FunnyOrOpposite to a person that enjoys 
              $TypeOfHumour humour.
\end{lstlisting}
\end{footnotesize}
\caption{Prompt for the vote's explanation for a particular personality.}
\label{prompt_explanation}
\end{figure}

Once each personality provides a vote and a reasoning for each joke, the next step is to audit those votes. Figure \ref{prompt_auditing} shows the crafted prompt to validate a vote. It is checking if ``the [Reasoning] explain why the [Joke] is [Classification]''. The LLM replies a simple ``yes" or ``no" answer. If the answer is a ``no", this vote is discarded. Other policies could be applied, for example, before discarding a vote, and the LLM could be re-prompted with different parameters in order to output a different explanation. Different personalities could be used as different auditors. Finally, different auditing questions/prompts could be used to validate each answer. We leave the study of those policies for future work.

\begin{figure}
\begin{lstlisting}[language=Python]
Provide an answer to the following [Question], replacing [Reasoning], [Joke] and [Classification] 
slots for their contents.

Question: Does the [Reasoning] explain why the [Joke] is [Classification]?

Joke: $JokeDescription
Reasoning: $ReasoningDescription
Classification: $ClassificationDescription

Answer:
\end{lstlisting}
\caption{Prompt for auditing vote's explanations.}
\label{prompt_auditing}
\end{figure}

\subsection{Score Aggregation}
The last step is to aggregate the validated votes into a single score that represents the crowd’s judgement. This can be done using different functions. In this paper, we sum up the binary votes (1 - funny, 0 - not funny) as shown in Equation \ref{equation_votes}, where $i$ is a joke and $j$ a personality. Since we have four personalities, the crowd score is in the range [0,4] for each joke.

\begin{equation}
\footnotesize
CrowdScore_{i} = \sum vote_{ij}
\label{equation_votes}
\end{equation}

\section{Results}\label{results}

In this section, we present the experimental results using the Crowd Score method applied to a dataset of jokes found on \cite{toplyn2022}. This dataset has been selected because it contains recent jokes, which were not included in the training dataset of our target model GPT-3, and these are non-trivial jokes that rely on common sense instead of wordplay. On top of that, this dataset is one of the few that also has a score/rating of funniness rated by human judges, so we can compare AI voters with human ones.

\subsection{Experimental Setup}\label{exp_setup}

The dataset in \cite{toplyn2022} is composed of 13 inputs (headlines). Each one was submitted to different human and non-human comedians, which generated each joke by completing the input. The comedians are: Open AI GPT-3, Witscript, Witscript 2 and Human \cite{toplyn2022}. GPT-3 is the LLM provided by OpenAI, the same one used in our experiments. Witscript and Witscript 2 are different versions of a joke writing algorithm developed by an expert comedian. And finally, the jokes were also generated by a professional human comedian. Four jokes were generated for each input, which is a total of 52 jokes. All jokes are within the aggressive/self-defeating spectrum, in contrast with the affiliative/self-enhancing one. The results in \cite{toplyn2022} show that jokes generated by GPT-3 achieved the lowest scores, while human jokes achieved the highest scores. 

For all experiments, we set GPT-3 with text-davinci-002, the temperature and the top P are set 1 in order to maximize creativity. All experiments were run 3 times and the results shown are an arithmetic average/mean. All code used for this research can be found at\footnote{https://github.com/creapar/crowdscore/}.

In order to evaluate the accuracy of the voting questions, we used two metrics: f-score and balanced accuracy. The positives and negatives correspond to if a joke is funny or not.

\subsection{Voting Question}

In this section, we discuss the results of the voting question step. First, all jokes from \cite{toplyn2022} were sorted by their corresponding human rating. Then, the four least funny (negatives) and the four funniest jokes (positives) were selected to compose a test dataset for finding the voting question with the highest accuracy in classifying jokes into funny and not funny. In this first set of experiments, the prompts in Figures \ref{prompt_voting_zero} and \ref{prompt_voting_few} were used. The positives were fixed in the word ``Funny" and the negatives (e.g opposite of funny) were varied between: ``Not funny", ``Dumb", ``Unfunny", ``Not Amusing", ``Sad", ``Serious", ``Dull" and ``Boring". Two versions were tested: zero-shot and few-shot. In the few-shot version, an example of a funny and another with a not funny joke were provided to the LLM.

Table \ref{table_voting_question_small} shows the results using the F-score and accuracy for this small dataset. As it can be observed, ``Boring" and ``Dull" presented the best accuracy, which means that they are better to split the jokes into funny and not funny. Few-shot prompting leads to better results than zero-shot as expected. However, picking the least appropriate opposite word can reduce the balanced accuracy by 26\% and 25\% for zero-shot and few-shot respectively. These results show that using a naive word such as ``Not Funny" can lead to a significant reduction in the accuracy.

\begin{table*}[th!]
\centering
\begin{tabular}{|l|ll|ll|}
\hline
\multicolumn{1}{|c|}{} & \multicolumn{2}{c|}{F-Score}                                  & \multicolumn{2}{c|}{Balanced Accuracy}                         \\ \hline
\multicolumn{1}{|c|}{} & \multicolumn{1}{c|}{Zero-Shot} & \multicolumn{1}{c|}{Few-Shot} & \multicolumn{1}{c|}{Zero-Shot} & \multicolumn{1}{c|}{Few-Shot} \\ \hline
Funny / Boring                 & \multicolumn{1}{l|}{0.89}      & 1                             & \multicolumn{1}{l|}{0.88}      & 1                             \\ \hline
Funny / Dull                   & \multicolumn{1}{l|}{0.89}      & 1                             & \multicolumn{1}{l|}{0.88}      & 1                             \\ \hline
Funny / Serious                & \multicolumn{1}{l|}{0.8}       & 0.8                           & \multicolumn{1}{l|}{0.75}      & 0.75                          \\ \hline
Funny / Sad                    & \multicolumn{1}{l|}{0.8}       & 0.8                           & \multicolumn{1}{l|}{0.75}      & 0.75                          \\ \hline
Funny / Not Amusing            & \multicolumn{1}{l|}{0.75}      & 0.86                          & \multicolumn{1}{l|}{0.75}      & 0.88                          \\ \hline
Funny / Unfunny                & \multicolumn{1}{l|}{0.67}      & 0.86                          & \multicolumn{1}{l|}{0.62}      & 0.88                          \\ \hline
Funny / Dumb                   & \multicolumn{1}{l|}{0.67}      & 0.86                          & \multicolumn{1}{l|}{0.62}      & 0.88                          \\ \hline
Funny / Not Funny              & \multicolumn{1}{l|}{0.67}      & 0.86                          & \multicolumn{1}{l|}{0.67}      & 0.88                          \\ \hline
\end{tabular}
\caption{Accuracy results using the small dataset composed of the four funniest jokes and the four least funny jokes for zero-shot and few-shot prompting of the voting question.}
\label{table_voting_question_small}
\end{table*}

Based on these results, we ran a full experiment with all 52 jokes just for Boring (the best accuracy) and Not Funny (naive version). In this case, we considered all jokes with human rating larger than or equal to 2 as funny, and the ones less than 2 as not funny. In this case, 15 jokes were considered not funny and 37 considered funny. Table \ref{table_voting_question_large} shows that for few-shot there is a small improvement when using ``Boring" instead of ``Not Funny". However, for the zero-shot version the difference in accuracy is up to 19\% regarding the F-score. This result shows that picking the correct opposite word can improve the classification accuracy.

\begin{table*}[t]
\centering
\begin{tabular}{|l|ll|ll|}
\hline
\multicolumn{1}{|c|}{} & \multicolumn{2}{c|}{F-Score}                                  & \multicolumn{2}{c|}{Balanced Accuracy}                         \\ \hline
\multicolumn{1}{|c|}{} & \multicolumn{1}{c|}{Zero-Shot} & \multicolumn{1}{c|}{Few-Shot} & \multicolumn{1}{c|}{Zero-Shot} & \multicolumn{1}{c|}{Few-Shot} \\ \hline
Funny/Boring           & \multicolumn{1}{l|}{0.78}      & 0.83                          & \multicolumn{1}{l|}{0.7}       & 0.78                          \\ \hline
Funny/Not Funny             & \multicolumn{1}{l|}{0.59}      & 0.81                          & \multicolumn{1}{l|}{0.6}       & 0.76                          \\ \hline
\end{tabular}
\caption{Accuracy results using the full dataset composed of 52 jokes for zero-shot and few-shot prompting of the voting question.}
\label{table_voting_question_large}
\end{table*}

\begin{figure}[th!]
\begin{center}
\includegraphics[scale=0.5]{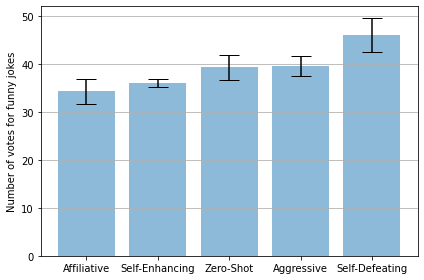}
\end{center}
\caption{Number of votes for funny jokes by each personality compared to the zero-shot version (no personality).}
\label{figure_personalities}
\end{figure}

\subsection{Personality Induction}
In order to induce personalities, we rely on the prompt in Figure \ref{prompt_personality}. The personalities are based on the description of the types of humour which are known by the LLM. We evaluated four types of humour: affiliative, self-enhancing, aggressive and self defeating. Since the dataset is composed of jokes in the spectrum of aggressive and self-defeating jokes, it is expected that the aggressive and self-defeating personalities find more jokes funny than the other two personalities as confirmed by Figure \ref{figure_personalities}. These personalities were induced using the zero-shot version, since the use of few-shot prompting (examples) overwrites the personality induction and all personalities achieve similar results. It is also important to note that the zero-shot version without personality induction lies in the middle of the spectrum. Zero-shot seems to be an average of all other types of humour, indicating that GPT-3 seems not to be biased towards a specific type of humour.

\begin{figure}[th]
\begin{center}
\includegraphics[scale=0.5]{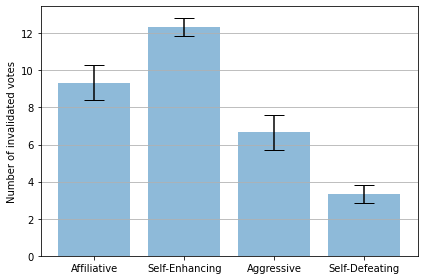}
\end{center}
\caption{Number of invalidated votes per personality.}
\label{figure_auditing}
\end{figure}

\begin{figure}[th]
\begin{center}
\includegraphics[scale=0.5]{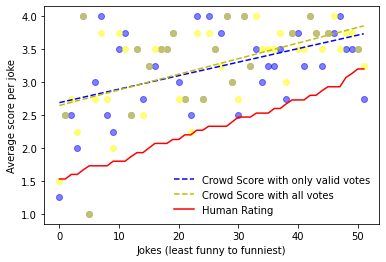}
\end{center}
\caption{Average scores from human judges from \cite{toplyn2022} and the average Crowd Score - all votes and only valid ones.}
\label{figure_crowd_score}
\end{figure}

\subsection{Auditing}

Reliability is an important issue in LLMs and can lead to inaccurate results. In order to tackle this issue, we introduced an auditing prompt as shown in Figure \ref{prompt_auditing} to ensure that each model's decision/classification is based on convincing explanations. Normally, a human expert would be needed to check those explanations, introducing a bottleneck in the jokes rating. The Crowd Score method assumes that no human intervention is needed in the whole process of assessing the creative artifact, so the auditing is also automated. 

Overall, the percentage of invalidated votes was 14\% of all votes. It means that 14\% of the explanations were not in accordance with the classification as Funny or Boring. Figure \ref{figure_auditing} shows the results for a number of invalidated votes per personality. This is explained by the fact that all jokes in the dataset are aggressive/self-defeating, so the content of the jokes provides more context (input) to the LLM, making it ``easier" to justify the classification of those jokes. In Figures \ref{example_positive} and \ref{example_negative} in the Appendix, we show examples of a valid and an invalid vote.

In future work, it would be beneficial to use another model to audit the explanations to avoid any bias from the model evaluating itself. This could also be alleviated by inducing different personalities in the auditing process.

\subsection{Score Aggregation}

The Crowd Score is calculated using Equation \ref{equation_votes}. In order to compare our results with \cite{toplyn2022}, where human judges rated jokes from 1 (not a joke) to 4 (very good joke), we 
 also normalized the crowd score in the range [1,4]. Figure~\ref{figure_crowd_score} shows the trending curves for the Crowd Score using all votes and only valid votes as compared with the human ratings. The results show that the Crowd Score is able to follow a similar tendency of the human evaluators, using all votes or only the valid ones. The scatter plot shows the scores for each joke. It is important to note that for the top 10 jokes, only one score in 20 was below 3, and for the bottom 10 jokes, only 5 in 20 scores above 3. This indicates that the Crowd Score is fairly accurate at detecting the funniest and the least funny jokes. 

The percentage of invalid votes is small enough not to affect the overall tendency. Interestingly, the curve for all votes is a bit more similar to the human rating. Inspection of the results by a human expert showed that the current auditing prompt is efficient to capture positives, but around 40\% of votes that it still invalidates should be considered valid.

\section{Conclusion} \label{conclusion}

In this paper, we present the Crowd Score, a new method for assessing the creativity of artefacts using LLMs as AI judges. We applied this method to assess the funniness of jokes from \cite{toplyn2022} in a crowd of four voters with different humour types: affiliative, self-enhancing, aggressive and self-defeating. Our results show that prompting LLMs for humour detection is a viable approach. The Crowd Score follows the same trend as human judges by assigning higher scores to jokes that are also considered funnier by human judges. This method could be applied to other creative domains such as story, poetry, slogans, etc. It could help both the adoption of a flexible and accurate standard approach to compare different work in the CC community under a common metric and by minimizing human participation in assessing creative artefacts. We can accelerate the prototyping of creative artefacts and reduce the cost of hiring human volunteers to rate creative artefacts.

The findings in this research point in the direction that AI voters can be used as judges of humour in the process of evaluating jokes. This would reduce the bottleneck of asking human judges to assess jokes. This could enable comedians to have an instantaneous feedback system to evaluate their jokes before releasing them to the public.

This research opens up many possibilities for future work. Firstly, automated ways of finding the best voting questions could be explored. Secondly, how much description is enough to induce a personality is also an open problem and it can depend on the target domain and public. Another open question is on how to audit LLM's explanations in order to achieve the highest reliability. This could be done by creating multiple prompts/questions verifying different aspects of the explanation, or by using external auditors through other LLMs. Auditors could also be configured to assume different personalities. Also, the Crowd Score itself could be aggregated in different ways. An important future work is to use AI voters as judges in other domains such as stories, slogans, poetry etc. to be part of a fully automated system to generate and evaluate those artifacts too. Finally, the Crowd Score makes it easier to reproduce and compare results from different papers in contrast to working with human as judges. This could accelerate the research in the Computational Creativity community.

\section*{Acknowledgments}
We would like to thank the University of Leicester for supporting this research, in particular the Computing and Mathematical Sciences Department.



\bibliographystyle{IEEEtran}  

\newpage
\section*{Appendix}
\begin{figure}[h]
\tiny
\begin{lstlisting}[language=Python]
Joke: A man was caught during a home burglary wearing only lingerie. He was charged with panty larceny.

Classification: Funny.

Reasoning: Both of these events are negative, or self-defeating, for the man involved. 
This makes the situation funny for someone who enjoys self-defeating humour.
\end{lstlisting}
\caption{Example of a valid vote for a self-defeating humour.}
\label{example_positive}
\end{figure}

\begin{figure}[h]
\begin{footnotesize}
\begin{lstlisting}[language=Python]
Joke: A truck was stolen that contained half a million dollars' worth of handbags. Truck handbags? Yep, of vanbags.

Classification: Funny.

Reasoning: First, the joke is not aggressive. It is about a truck that was stolen and the truck contained handbags. 
Second, the joke is not particularly funny. It is simply a statement about a truck that was stolen. 
Therefore, the joke is Boring for a person that enjoys aggressive humour.
\end{lstlisting}
\caption{Example of an invalid vote for an aggressive type of humour.}
\label{example_negative}
\end{footnotesize}
\end{figure}
\end{document}